# Functional advantages of an adaptive Theory of Mind for robotics: a review of current architectures

Francesca Bianco
School of Computer Science and
Electronic Engineering
University of Essex
Colchester, United Kingdom
fb18599@essex.ac.uk

Dimitri Ognibene
School of Computer Science and
Electronic Engineering
University of Essex
Colchester, United Kingdom
dimitri.ognibene@essex.ac.uk

*Abstract*— **Great advancements have been achieved in the field of robotics, however, main challenges remain, including building robots with an adaptive Theory of Mind (ToM). In the present paper, seven current robotic architectures for human-robot interactions were described as well as four main functional advantages of equipping robots with an adaptive ToM. The aim of the present paper was to determine in which way and how often ToM features are integrated in the architectures analyzed, and if they provide robots with the associated functional advantages. Our assessment shows that different methods are used to implement ToM features in robotic architectures. Furthermore, while a ToM for false-belief understanding and tracking is often built in social robotic architectures, a ToM for proactivity, active perception and learning is less common. Nonetheless, progresses towards better adaptive ToM features in robots are warranted to provide them with full access to the advantages of having a ToM resembling that of humans.**

## I. INTRODUCTION

In the last decades, we have assisted to an incredible advancement in the field of robotics. Increasingly sophisticated robots with complex abilities and behaviors have been developed and research has been conducted to permit their application in everyday scenarios [1-3]. However, although the humanoid aspect of robots has increased humans' positive attitude towards them, their still limited social capabilities have a negative impact on humans' trust and acceptance of robots as social companions (uncanny valley effect) [4]. Therefore, several clever robotic architectures have been created to equip robots with human-like social skills and improve human-robot interactions. While some of these architectures only aimed at equipping robots with social skills for an effortless interaction with humans [5], others were inspired by our knowledge of human social understanding providing plausible models of human cognition [6, 7]. Nonetheless, building robots with a Theory of Mind (ToM), thus with the ability to understand others' intentions, beliefs and desires [8], remains amongst the current challenges of social interactive robotics [9]. For example, although robots are able to understand agents' actions and learn plans, they do not appreciate the actual final goal of an action, i.e., the underlying intention [10]. In addition to understanding the reasons behind agents' actions, inferring such hidden states represents an invaluable advantage when predicting their future behavior [9]. Therefore, ToM reasoning in robots is expected to ensure optimal human-robot interactions [11]. In section II of the present paper, some of the most current architectures for social robots will be described. Successively, the functional advantages of a ToM for robotics will be illustrated in section III. Finally, the identified robotic architectures will be discussed in light of the mentioned functional advantages of ToM and the future challenges for building robots as social agents will be highlighted.

## II. SOCIAL ROBOTIC ARCHITECTURES

In this section, some of the main robotic architectures developed to equip robots with social skills will be briefly described.

The BASSIS (biomimetic architecture for situated social intelligence systems) architecture is presented by Petit et al. [10]. It aims at building robots able to learn in real-time and adapt a shared plan through their interaction with the environment, including humans, to perform successful cooperative, turn-taking tasks. This architecture utilizes multiple modalities to ensure online learning, including visual perception to imitate actions performed by the human, proprioception of the physical manipulation of the robot's arm by the human instructor to teach a new movement, and auditory perception to understand language commands. Finally, the authors showed this architecture to perform effectively both on the NAO and iCub humanoid robots in the "clean the table" and "hide a toy under a box" cooperative tasks, respectively.

The architecture proposed by Görür et al. [7] integrates a ToM model into robots' decision-making process during collaborative tasks. The aim of this addition is to allow robots to infer humans' intentional and emotional states during shared plan execution and adapt to changes of such internal states to behave less intrusively and more appropriately. Similarly to BASSIS, this architecture also relies on multi-modal estimation of humans' mental states. More specifically, it employs visual perception, including humans' location, pose, gaze, visual perspective, facial emotion expressions, and auditory perception for command recognition. Furthermore, a Hidden Markov Model (HMM) is used to estimate the action performed by the interacting agent. An example implementation is given in the paper which describes a robot able to model a human's desire for an object and the process of acquiring it. While the robot first defines some conditions related to the beliefs of the agent, it also uses visual information to determine whether the human is performing actions related to the desired object and his intentions with regards to that object.

Similarly, a ToM model is integrated in the framework illustrated in Devin et al. [12] which permits the estimation of both the state of the environment and human partners' internal states, in particular goals and plans, which are considered by the robot for successful human-robot shared plans performance. More specifically, this architecture responses to unexpected situations and ambiguity which may happen during human-robot interaction given humans' initiative. Therefore, it aims to permit robots to adapt to humans' changing decisions and to provide only necessary information to their human partners, without being intrusive, achieving a fluent execution of a shared plan. Also this architecture relies on multi-modal perception to infer mental states and achieve a successful collaborative task. Specifically, visual perspective-taking is provided to robots, which allows them to view the world from the perspective of their human partner, together with the recognition of basic vocal commands. Furthermore, this architecture consists of two planner modules which enable the robot to search for the optimal plan to complete the shared goal, considering human-aware costs, their safety and comfort. Two implementations of this architecture were presented in this paper, i.e., the "clean the table" and "inventory scenario" collaborative tasks. The robots were shown to successfully complete the tasks with their human partner by providing the latter with only the necessary information and by reducing their intrusive behavior.

The robotic architecture presented by Lemaignan et al. [5] also enables robots to infer the mental states (i.e., beliefs, intensions, and desires) of the agents they are interacting with. However, it expands on the architectures previously mentioned as it comprises a deliberative layer which runs other "background deliberative tasks" (i.e., situation assessment, action monitoring and processing of non-imperative speech) that are not automatically activated by agents' mental states. They enable perspective taking and permit an effective human-robot interaction together with an increasingly proactive behavior of robots during such interactions. Similarly to the other architectures, this framework relies on multi-modal disambiguation of the agents' behavior. In fact, both the visual perception (i.e., visual and spatial perspective taking, recognition of gaze direction and gestures) and the auditory perception (i.e., verbal interaction) are utilized. An implementation example is provided in the paper, in which the "cleaning the table" cooperative task between a robot and a human is illustrated. This architecture was shown to allow the robots to reason about the human partner's mental states and generate, monitor and take part to human-robot shared plans.

Another approach was instead taken by Rabinowitz et al. [13] who employed a ToM neural network, ToMnet, to infer agents' mental states (i.e., beliefs, intentions and desires) online. Specifically, ToMnet utilizes a meta-learning approach, i.e., it models the behavior of an agent and predicts his mental states through a very small number of observations by applying strong prior models learnt through the interaction with other agents and adapting them to the current state of the agent observed. In contrast to the previous examples, this ToMnet exclusively relies on the visual modality to have access to agents' mental states. In fact, their past and current trajectories are considered and fed into a prediction net, which outputs predictions of future behaviors. In their paper, Rabinowitz et al. demonstrate the capacity of this ToMnet to learn rich models of other agents inferring their mental states. The ability to recognize agents' false belief is provided through both the "food-truck" task (in which agents are observed to follow different trajectories to several food-trucks based on their desires and beliefs) and the Sally-Anne task (which is a standard paradigm utilized in human studies to determine people's ToM abilities [14]).

Another related issue is assessed in Vanderelst & Winfield [15], who created an architecture for "ethical" robots that takes into account agents' mental states. Briefly, this architecture includes an ethical layer in which the simulation principle is utilized to simulate the motor and sensory consequences of an agent's different behaviors and the resulting internal states. Finally, the robot controller selects the best behavior alternative, based on the action desirability of each simulated behavior and emotional state. This architecture relies on multi-modal perception to determine agents' behaviors, including the visual (i.e., gaze direction recognition and target location) and auditory (i.e., spoken commands) modalities. In this paper, this ethical architecture was implemented on a NAO humanoid robot which interacted with another NAO robot representing a human partner. Through four tasks inspired by Asimov's Laws of robotics (i.e., self-preservation, obedience, human safety, and human safety and obedience), the ethical NAO robot was shown to successfully prevent the NAO robot used as a proxy for the human partner from coming to harm.

Finally, the HAMMER (Hierarchical Attentive Multiple Models for Execution and Recognition) architecture will be described. Demiris & Khadhouri presented in their paper [6] this computational architecture that allows robots to select and execute an action, as well as understand it when shown by a demonstrator (thus predicting his future behavior). Similarly to the previous framework, this architecture is inspired by the simulation hypothesis. Specifically, it relies on the collaboration between inverse models (which assess the state of the system and the desired goal, and output motor commands to achieve that goal) and forward models (which simulate the different sensory consequences resulting from the execution of such commands) to understand agents' intentions and predict their behavior. This architecture is based on the visual perception of agents' movements, which is controlled by top-down signals to orient the robot's attention towards those information necessary to confirm its hypothesis concerning the demonstrator's action. This architecture was implemented on a robot that conducted an action recognition task while observing a human demonstrator performing an object-oriented action. The robot successfully performed the task and the attentional mechanism acting over the inverse models was suggested to reduce robots' computational costs.

III. FUNCTIONAL ADVANTAGES OF TOM FOR ROBOTICS

The ToM ability is present in humans since an early age and represents an essential skill for the successful understanding of other agents' behavior and mental states [16, 8]. Several attempts of integrating a ToM model into robotic architectures exist in the literature, however, as mentioned earlier, creating robots with a ToM remains a big challenge for the field of robotics. In the next paragraphs, the main functional advantages, from us identified, of building robots with an adaptive ToM will be briefly described with the aim of inspiring future studies in this direction.

*A. A ToM for coordinating and managing false beliefs*

Beliefs are generally included amongst humans' mental states [8] and their understanding and tracking is considered an essential requirement for successful human-robot interactions [12], especially during collaborative tasks. The realization that other people represent the world from their perspective, and thus may have similar or different beliefs, is a distinguishing feature of the ToM (or mentalistic) account [16]. Therefore, equipping robots with a perspective taking capability may be a resourceful way to improve their correct inference of human mental states and ameliorate their social interactions. False beliefs tasks are generally considered the standard experimental paradigm to assess belief understanding and ToM abilities in humans (see [14] for further details). Therefore, enabling robots to solve such tasks may represent a great step towards more effortless and fluid human-robot interactions. This would mean systems equipped with an adaptive ToM, where the programmer's input is little and where robots are able to autonomously attribute mental states, reason about them and appropriately react to them.

*B. A ToM for proactivity and preparation*

Providing robots with an adaptive ToM would also result in robots assuming a proactive behavior during social tasks, improving human-robot interactions. Being able to determine other agents' beliefs, desires and intentions may in fact allow the anticipation of their behavior prior to any concrete action. Specifically, proactivity implies a lower reliance on bottom-up inputs in favor of additional top-down control on the robot's response in a given situation. This characteristic is important for the successful application of robots in everyday social settings and collaborative tasks, as social contexts are highly dynamic and robots are required to act prior to an event rather than just to react to it. Furthermore, the top-down nature of ToM would provide a substantial advantage during the preparation for interactions with other agents. Inferring and anticipating agents' beliefs and intentions would in fact allow robots to prepare for an efficient and fluid interaction (e.g., positioning themselves in a position easy for other agents to spot).

*C. A ToM for (active) perception*

Associating intentions and mental states to agents' behavior may encourage the observer to actively search for cues, such as unnoticed affordances, to acquire a better understanding of a given situation and enable more precise predictions [17]. Active perception may be necessary to eliminate the passive nature of robots' exploration and understanding of the environment and agents, which is in contrast with the ecological behavior seen in humans and limits the quality of human-robot interactions.

*D. A ToM for learning*

An adaptive ToM for learning would imply a different way of learning about the world, resolving some robotics challenges identified by Lake et al. [18]. Specifically, most DNN-based action recognition systems do not currently differentiate between the learning of passive objects dynamics (e.g., the movement of a ball) from that of agents' behavior (e.g., entering different rooms to find a desired object while searching), oversighting the intentionality and belief state that marks humans' behaviors. In particular, when learning about others' behaviors, subtasks such as searching and correcting disturbances would consist in complex and ambiguous training examples. In contrast, we propose that equipping robots with mental states understanding and contextualization would provide a means to distinguish between passive object dynamics and agents behaviors. Thus, integrating ToM development principles in the blueprint of an adapting neural architecture for social interaction may result in a more time- and cost-efficient learning process compared to current DNNs [6]. This will also decrease the size of the datasets needed to train robots, reducing the effect of human errors involved in dataset preparation and permitting increased accuracy.

## IV. DISCUSSION

Whether the functional advantages identified in the previous section are taken into account by the architectures described in section II will now be highlighted. A summary of these findings is provided in TABLE I below.

TABLE I.   ToM FUNCTIONAL ADVANTAGES INTEGRATED IN CURRENT ROBOTIC ARCHITECTURES

| Robotic Architectures | ToM Functional Advantages | | | |
|---|---|---|---|---|
| | *False beliefs* | *Proactivity* | *Active Perception* | *Learning* |
| Petit et al. [10] | X | X | X | X |
| Görür et al. [7] | ✓ | X | X | ✓ |
| Devin et al. [12] | ✓ | X | X | ✓ |
| Lemaignan et al. [5] | ✓ | ✓ | X | X |
| Rabinowitz et al. [13] | ✓ | X | X | X |
| Vanderelst & Winfield [15] | ✓ | X | X | ✓ |
| Demiris & Khadhouri [6] | ✓ | ✓ | X | ✓ |

*A. A ToM for coordinating and managing false beliefs*

Several papers integrated the ability of visual perspective taking in their architecture [7, 12, 5] to infer the mental states (thus beliefs) of agents. In fact, by enabling robots to put themselves in the agents' shoes and infer their sensorial access, a better recognition of mental states and increased performances in belief recognition tasks were observed. Future studies should however create innovative adaptive ToM architectures which also aim at equipping robots with mental perspective taking. This would in fact signify that, in addition to mental states and beliefs which can be inferred based on geometrical principles, robots would also be able to share and have full access to all the mental states that an agent may possess, such as emotions, resulting in improved human-robot interactions. Another approach was instead presented by Rabinowitz et al. [13], who do not rely on perspective taking to understand agents' mental states but proposed a NN able to predict the behavior of multiple agents in a false-belief situation given their past and current trajectories. In contrast, papers [15, 6] utilized the simulation account to allow robots to infer the mental states and predict the agents' behavior. In contrast, paper [10] describes the only architecture which does not consider belief understanding and tracking and focuses on the learning of shared plans instead.

*B. A ToM for proactivity and preparation*

The only paper that discusses robots' proactive behavior during social interactions is [5]. The authors indicate that one of the main roles of the SHARY module, part of their architecture, is to control the production and execution of shared plans, through a refinement based on context and specific agents. They suggest that, this way, the robots are able to adapt to different environments and levels of involvement during tasks, thus providing a proactive behavior. The HAMMER architecture [6] may also be considered to equip robots with a proactive behavior, given the top-down control over the attentional system which allows robots to only respond to stimuli necessary for the identification of the correct future behaviors. However, as we have seen in the previous paragraph, this architecture is limited to the understanding of agents' intentions. Finally, [7, 12] provide some advancements related to the preparation of a robot to social interactions. More specifically, in [7] robots take into account the human's potential unwillingness to be assisted from the robot, whereas in [12] robots only execute their actions when they are considered as needed and possible, thus rendering the human-robot interactions more fluid and effortless.

*C. A ToM for (active) perception*

None of the architectures here described seem to take this feature into account. To provide an example, the Situation Assessment Module in [12], which has the role of processing sensory data and maintain the state of the world (and agents), is not able to compute non-observable information and facts. The only way for the robot to access such non-observable information is through a dialog with the agent or the observation of actions which have these information in their side-effects. Another example of passive perception is provided by [5], whose architecture makes the assumption that the robot benefits of a nearly perfect perception and that objects are already known in advance. Similarly, [13] assume the observer, thus the robot, to have access to the states and actions of all agents, which is unlikely in real-life interactions. A related issue, i.e., sensory prioritization or "visual attention", is provided in HAMMER [6], however it does not deal with missing information (occluded or out of the field of view) and does not focus on counterfactual cues, as its policy always observes the most likely candidate cues. Although an information theoretic extension of HAMMER able to address this issues was presented in Ognibene & Demiris [19], its action set is fixed and limited to reaching actions.

*D. A ToM for learning*

The architecture proposed by [7] presents a system able to learn the preferences of an agent and plan during a shared plan task. By taking agents' emotions and intentions into account, this architecture provides a means of distinguishing the learning of agents' behavior from that of passive object dynamics. The ToM Manager in [12] estimates and maintains the mental states of agents involved in a cooperation task based on human-aware costs, thus also enabling to learn specifically about humans' behavior. In contrast, as we have previously mentioned, papers [15, 6] follow the simulation principle for inferring mental states to agents, thus robots are able to specifically learn about the agents' behavior and distinguish it from other passive object dynamics. The architecture described in [13] utilizes the meta-learning method to infer intentions, which is based on the acquisition of priors about the common behaviors of agents in the training set and then on a refinement based on the observation of specific agents. However, this method does not seem to rely on a ToM to learn about the agents' behavior. In fact, this ToMnet recognizes behaviors from pure observation and this approach can be broadened to passive object dynamics; thus, it is not specific to humans' behavior. Finally, architecture [5] is based on a large set of high-level actions for prototyping and therefore heavily relies on datasets. Furthermore, this architecture does not seem to properly distinguish the learning between humans and objects, as it mainly focuses on geometric features and affordances. In addition, it requires programmers to manually label concepts with human terminology for verbal interactions.

## V. CONCLUSION

To summarize, four main functional advantages of a ToM for robotics were here identified and discussed in relation to seven current robotic architectures. While it appears that coordinating and managing false-beliefs has been generally addressed, the other functional advantages have been infrequently or separately integrated in the analyzed architectures. We can conclude that building robots with an adaptive ToM will likely further enhance robots' capabilities and improve human-robot interactions, by: (a) enabling robots to take the agents' mental perspective for false-belief understanding, (b) providing robots with proactivity and preparation for more fluid and effortless social interactions, (c) supporting active perception to more closely replicate humans' behavior during interactions, and (d) providing a means to distinguish between the learning of humans' behavior from that of passive object dynamics as well as reducing the need of datasets for the inference of mental states.